\documentclass[conference]{IEEEtran}
\ifCLASSINFOpdf
\else
\fi

\usepackage{microtype}
\usepackage{graphicx}
\usepackage{subfigure}
\usepackage{amsmath}
\usepackage{booktabs} 
\usepackage{url}
\usepackage{xcolor}

\newcommand{\specialcell}[2][c]{%
  \begin{tabular}[#1]{@{}c@{}}#2\end{tabular}}

\usepackage{hyperref}

\usepackage{algorithm}                                \usepackage{algorithmic} 

\newcommand{\DS}{\mathcal{D}}

\usepackage{hyperref}

\makeatletter
\newcommand{\printfnsymbol}[1]{%
  \textsuperscript{\@fnsymbol{#1}}%
}
\makeatother

\usepackage{cite}
\usepackage{amsmath,amssymb,amsfonts}
\usepackage{algorithmic}
\usepackage{graphicx}
\usepackage{textcomp}
\usepackage{xcolor}
\usepackage{eqparbox}
\begin{document}

\title{Variational Resampling Based Assessment of Deep Neural Networks under Distribution Shift
	\thanks{Equal contribution with alphabetical order}
}

\author{
	\IEEEauthorblockN{1\textsuperscript{st} Xudong Sun\IEEEauthorrefmark{1}, 1\textsuperscript{st} Alexej Gossmann\IEEEauthorrefmark{2}, 3\textsuperscript{nd} Yu Wang\IEEEauthorrefmark{3}, and 4\textsuperscript{th} Bernd Bischl\IEEEauthorrefmark{4}}
	\IEEEauthorblockA{\IEEEauthorrefmark{1}
		\textit{Department of Statistics} \\
		\textit{Ludwig Maximillian University of Munich}, Munich, Germany \\
		smilesun.east@gmail.com}
	\IEEEauthorblockA{\IEEEauthorrefmark{2}
		\textit{Center for Devices and Radiological Health} \\
		\textit{U.S. Food and Drug Administration}, Silver Spring, MD, USA \\
		alexej.gossmann@fda.hhs.gov
	}
	\IEEEauthorblockA{\IEEEauthorrefmark{3}
	\textit{Technical University of Munich}, Munich, Germany}
	\IEEEauthorblockA{\IEEEauthorrefmark{4}
		\textit{Department of Statistics} \\
		\textit{Ludwig Maximillian University of Munich}, Munich, Germany \\
		bernd.bischl@stat.uni-muenchen.de}
}

\maketitle

\begin{abstract}
	A novel variational inference based resampling framework is proposed to evaluate the robustness and generalization capability of deep learning models with respect to distribution shift.
We use Auto Encoding Variational Bayes to find a latent representation of the data, on which a Variational Gaussian Mixture Model is applied to deliberately create distribution shift by dividing the dataset into different clusters. Wasserstein distance is used to characterize the extent of distribution shift between the generated data splits. We compare several popular Convolutional Neural Network (CNN) architectures and Bayesian CNN models for image classification on the Fashion-MNIST dataset, to assess their robustness and generalization behavior under the deliberately created distribution shift, as well as under random Cross Validation. Our method of creating artificial domain splits of a single dataset can also be used to establish novel model selection criteria and assessment tools in machine learning, as well as benchmark methods for domain adaptation and domain generalization approaches.
\end{abstract}


\IEEEpeerreviewmaketitle

\section{Introduction}\label{intro}

Recent studies have shown that deep learning methods may not generalize well beyond the training data distribution.
For instance, deep learning models are vulnerable to adversarial perturbations \cite{carlini2017towards}, are prone to biases and unfairness \cite{Chen2018-gp}, or may significantly but unknowingly depend on confounding variables resulting from the training data collection process \cite{Badgeley2019-be}.
In this work we focus on distribution shift, which is another important phenomenon that can have a significant negative impact on the performance of deep learning models \cite{wen2014robust}.
Addressing the problems related to distribution shift is especially crucial for medical applications of machine learning \cite{sun2019high,Nestor2018-gk}, and other high-risk application areas.

Areas of machine learning research related to distribution shift include Transfer learning, which is the process of improving the predictive performance on a target domain by using related information from the source domain \cite{weiss2016survey}. Domain Adaptation adapts the source domain distribution to the target domain distribution to improve the performance of a target learner in transfer learning . In contrast to domain adaptation, Domain Generalization is the process of utilizing data from several domains to train a system that will generalize to previously unseen domains \cite{akuzawa2019adversarial}.

Although several domain adaptation and domain generalization benchmark datasets exist \cite{long2017deep}, they are either curated by human experts as combinations of multiple datasets with distribution shifts available a priori, or obtained through specific data manipulation techniques such as rotations \cite{akuzawa2019adversarial}. Therefore, these datasets depend on domain knowledge and are restricted to specific tasks. For new applications, collecting datasets with distribution shift for evaluation of algorithms may be expensive or even intractable.
Thus, to facilitate the study of distribution shift, as well as domain adaptation and domain generalization, there is a need for a general and efficient method to create benchmark datasets for evaluation of these approaches.

Furthermore, the robustness of machine learning models to distribution shift between subsets of the same dataset or subdomains of a single domain seems to not have been studied to a sufficient extent in the past.
For the various practical applications of machine learning these subsets or subdomains may also represent different sources of data, subpopulations within the target population, as well as other types of stratification, and the variability in performance of machine learning systems between them is often not considered although it is substantial in many cases.

While Cross Validation is a widely used resampling technique for model selection in Machine Learning \cite{guyon2010model,bommert2019benchmark} which generates random splits on a datasets, a resampling technique that can generate splits with distribution shift to evaluate a machine learning model does not seem to be known.
Thus, inspired by the resampling technique used in Restrictive Federated Model Selection over shifted distribution \cite{sun2019high} we propose a resampling technique to artificially create pseudo subdomains, which can serve as a benchmark method to evaluate distribution shift related problems and potential solutions.

Specifically, in this work we are interested in characterizing changes in model test performance under distribution shift over different subsets of the same dataset, i.e., when the feature distribution (but not the label distribution) of the test data is shifted relative to the distribution of features in the training data although both are subsets of the same dataset.

Our major contributions are:

\begin{itemize}
\item We propose a resampling framework, operating on a given dataset, which creates splits corresponding to different distributions by estimating the pseudo domain label of each instance using variational inference. We use the Wasserstein distance to quantify the amount of distribution shift created by our method. Our method can be used for the creation of benchmark datasets for domain adaptation and domain generalization.
\item We assess the robustness and generalization behavior of several image classification CNN architectures, including their corresponding Bayesian versions, under the distribution shift artificially created by our resampling method. Comparison with random Cross Validation shows our method can efficiently create distribution shift. The proposed resampling approach can be used to aid model selection.
\end{itemize}

\section{Prerequisite}\label{sec:prereq}

\textbf{Auto-encoding Variational Bayes, and Variational Autoencoder (VAE) \cite{kingma2013auto}:}\, To model the likelihood of data $p(x)$, a latent variable model, characterized by the posterior $p(z|x)$, is approximated by a variational posterior distribution $q_{\phi}(z|x)$, where the latent variable in the variational posterior is reparameterized as $z = g(\epsilon, x)$ with an auxiliary random variable $\epsilon \sim p(\epsilon)$ following an appropriate distribution. The conditional distribution $p_{\theta}(x|z)$ can be modeled as a Gaussian distribution with mean and variance parameters computed from $z$ by a decoder neural network. Evidence lower bound (ELBO) of the likelihood is optimized with respect to $\theta$ and $\phi$ using stochastic gradient descent (SGD) over the Monte Carlo estimation. As in an Auto Encoder \cite{jiulong2017detecting} the auxiliary variable $z$ also serves as a latent representation of an instance.
 
\textbf{Bayesian Neural Network \cite{blundell2015weight}:}\, Different from generative representation learning methods such as VAE, which model a variational approximator to the model posterior on the hidden units, a Bayesian Neural Network builds variational models on the weights of the network, which can also be used for exploration in Reinforcement Learning \cite{blundell2015weight,sun2019tutorial}. The negative of ELBO, which is the variational free energy $F(D,\phi)$, is optimized. In particular, $F(D,\phi) = KL(q(w|\phi)||p(w)) - E_{q}\left[\log p(D \mid w)\right]$, where $D$ is the data, $w$ is the vector of weights, $\phi$ represents the variational mean and variance parameters for the weight distribution, $KL(q(w|\phi)||p(w))$ is the KL-divergence between the prior distribution $p(w)$ and the variational posterior $q_{\phi}(w|D)$, and $E_{q}\left[\log p(D \mid w)\right]$ is the expectation of log-likelihood under the distributional distribution. With the variational free energy $F(D, \phi)$ as loss function, where weights $w$ are implicitly represented by $\phi$, backpropagation with respect to the weights could be translated to variational parameter. Like Auto-encoding Variational Bayes, Bayes by Backprop \cite{blundell2015weight} also starts from an independent noise distribution, but instead of transforming the noise together with observation data to latent units, Bayes By Backprop associates each weight with a variational mean and scale parameter to mix with the noise. 
Bayesian Convolutional Neural Network with Variational Inference (Bayesian CNN) \cite{shridhar2019comprehensive} extends the Bayes By Backprop approach \cite{blundell2015weight} to CNNs and utilize the local reparameterization trick \cite{Kingma2015-qg,molchanov2017variational,gal2015bayesian} which we will elaborate in the method section.
 
\textbf{Variational Gaussian Mixture Model:}\,
Variational Learning of Gaussian Mixture Models (VGMM) \cite{vgmm} uses joint Normal-Wishart distributions for the means and inverse covariance matrices in a mixture of Gaussians, and a Dirichlet distribution for the mixing parameters. Instead of a point estimate of the mean vector, VGMM uses a Normal distribution characterized by the hypermean. VGMM result in a superior data estimation compared to simple Gaussian Mixtures.

\textbf{Distribution Shift:}\, Let the random vector $x$ represent the features and let the random variable $y$ be the class label. In this work, we investigate the conditional distribution shift over $p(x|y)$ between datasets (e.g., between the training and the test data), while the marginal distribution $p(y)$ is shared across all datasets (cf. \cite{zhang2013domain}).

\textbf{Wasserstein Distance:}\, Wasserstein distance between two distributions $p_x$ and $p_y$ can be defined as $\phi_W(p_x,p_y)=\underset{\gamma \in \prod(p_x,p_y)}{\inf}\underset{(x,y)\sim\gamma(x,y)}{E}[c(x,y)]$ \cite{pmlr-v48-peyre16,memoli2011gromov,flamary2017pot}, where $\gamma\in \prod(p_x,p_y)$ is the transportation plan or joint distribution of $(x,y)$, with marginal distributions $p_x$ and $p_y$ respectively, and $c(x, y)$ is the cost of moving $x$ to $y$. The Wasserstein distance is calculated by taking the infimum with respect to the transportation plan $\gamma \in \prod(p_x,p_y)$. Wasserstein distance can be approximated to optimize Generative Adversarial Networks \cite{arjovsky2017wasserstein}. Compared to KL divergence, Wasserstein distance experiences no numerical problems even when the two distributions have no overlap. Hence, we use the Wasserstein distance to measure the distribution shift between two subsets of data on the latent space.

\textbf{t-SNE:\,} Stochastic Neighborhood Embedding \cite{hinton2003stochastic} uses a Gaussian density to model the conditional similarity between two points in a high dimensional space and a corresponding low dimensional embedding. The KL-divergence between the conditional similarity distributions is used as objective and is optimized with stochastic gradient descent. The t-SNE algorithm \cite{maaten2008visualizing} extends the conditional similarity to a symmetric version by adding the conditional similarity of both directions. Furthermore, it uses a Student-t distribution instead of a Gaussian distribution in the embedding.

\section{Methods}\label{sec:methods}

\subsection{Motivation}
\begin{figure}
	\centering
	\includegraphics[scale=1]{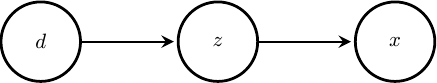}
	\caption{Data generation process from domains}
	\label{fig:dgp}
\end{figure}

Image data from different sources can come from different distributions, even when the same data collection process is meticulously followed \cite{Recht2019-cl}.
This phenomenon can be modeled by a directed probabilistic graphical model \cite{koller2009probabilistic} shown in Figure \ref{fig:dgp}, where the source or domain label $d$ generates the latent representation $z$, which further generates the observed image $x$. Given an observation $x$, we infer the latent representation $z$ and the domain label $d$ as described below and given in Algorithms \ref{alg:dscv-vae} and \ref{alg:merge}.
%
The proposed resampling method can be regarded as a worst case analysis aiming to identify the largest possible distribution shifts that can occur when a single dataset is split into multiple folds.

\subsection{VGMM-VAE-CV resampling scheme}

In the deep learning field a given dataset $\DS$ is typically split into disjoint subsets as $\DS = D_{train} \bigcup D_{val} \bigcup D_{test}$, where the training dataset $D_{train}$ is used for model training, the validation dataset $D_{val}$ aids with model selection (e.g., along the epochs), and the test dataset $D_{test}$ is used to evaluate the performance of the final model.
Many deep learning papers benchmark the performance of different models relying on the train-test split provided along with the dataset.

Another popular approach is $k$-fold cross validation, which splits the data randomly into $k$ disjoint subsets (folds or splits) of equal size, and which therefore should result in subsets with the same distribution.
Similarly to $k$-fold cross validation in this work we split the data into $k$ subsets. 
However, we aim to split the dataset such that each subset follows a different conditional distribution $p(x|y)$ of the features $x$ given a label $y$, as discussed in the following.

Since high dimensional clustering is challenging, we first train a representation of the dataset using a VAE (see Section \ref{sec:prereq}).
Instead of the observed distribution of $x$, for clustering we use the distribution of the latent space representation $z$ of $x$, denoted by $q_{\phi}(z|x)$.
We apply a VGMM (see Section \ref{sec:prereq}) on the latent representation space to assign each instance to a cluster.
When we cluster within subsets defined by each class label $y$ separately, our procedure corresponds to the conditional distribution shift, which refers to a change in $p(x|y)$, while $p(y)$ remains shared among the clusters.

Among the resulting clusters, one is used for testing and the remaining clusters are used for training and validation. We use random splitting to form the training and validation sets, which therefore share the same distribution. Analogously to conventional cross validation, the train-test process is repeated with each cluster playing the role of the test dataset once. The combination of assignments to $(D_{train} \bigcup D_{val}) \bigcup D_{test}$ yields a variation estimate on how good a model could generalize to $D_{test}$, similar to cross validation.
However, compared to conventional cross validation, our method provides a way to characterize the deep learning model's ability to generalize under distribution shift.

The whole process is summarized in Algorithm \ref{alg:dscv-vae}, along with Algorithm \ref{alg:merge}. Algorithm \ref{alg:merge} has an input argument $repeat$, which we set to $1$ in Algorithm \ref{alg:dscv-vae} for simplicity. We also only use $repeat=1$ in the experiments in this work due to the limited computational resources available to us. In larger scale benchmarks the modification $repeat = m > 1$ in our method would be analogous to how simple $k$-fold cross validation compares to $m$-times repeated $k$-fold cross validation.

\renewcommand\algorithmiccomment[1]{%
	\hfill\#\ \eqparbox{COMMENT}{#1}%
}
\begin{algorithm}[tb]
   \caption{VGMM-VAE-CV}
   \label{alg:dscv-vae}
\begin{algorithmic}
   \STATE {\bfseries Input:} dataset $\mathcal{D}$, number of classes $C$, number of folds $K$
   \FOR{$c=1$ {\bfseries to} $C$}
   \STATE $z^c = vae(D^c)$  \COMMENT{train VAE on data belonging to class $c$}
   \STATE $d^c_1,..., d^c_K = vgmm(z^c)$ 
   \ENDFOR
   \STATE $D_1, ..., D_K = merge(\{d_k^c\},\, repeat = 1)$
   \FOR{$k=1$ {\bfseries to} $K$}
   \STATE $D_{test} = D_k$
   \STATE $TrainValSet = {D} \setminus D_k$
   \STATE $D_{train}, D_{val} = randomSplit(TrainValSet)$
   \STATE $m_k = model\_init()$
   \FOR{$epoch = 1, 2, \ldots$}
      \STATE $PerfTrain_k, m_k = train(D_{train}, m_k)$ 
      \STATE $PerfVal_k = test(D_{val}, m_k)$
   \ENDFOR
   \STATE $PerfTest_k = test(D_{test}, m_k)$
   \ENDFOR
\end{algorithmic}
\end{algorithm}
\begin{algorithm}[tb]
   \caption{merge}
   \label{alg:merge}
\begin{algorithmic}
   \STATE {\bfseries Input:} $repeat \in \{1, 2, \ldots\}$, $d_k^c$ for $c = 1$ {\bfseries to} $C$ and $k = 1$ {\bfseries to} $K$
   \FOR{$i=1$ {\bfseries to} $repeat$}
   \IF{$repeat > 1$}
   \FOR{$c=1$ {\bfseries to} $C$}
   \STATE $d^c_1, ..., d^c_K$ = Shuffle($d^c_1, ..., d^c_K$)
   \ENDFOR
   \ENDIF
   \FOR{$k=1$ {\bfseries to} $K$}
   \STATE $D_k = \bigcup_{c=1}^C d_k^c$
   \ENDFOR
   \STATE yield $D_1, D_2, ..., D_K$
   \ENDFOR
\end{algorithmic}
\end{algorithm}

\subsection{Architecture of the VAE}
Following \cite{chen2016infogan}, given a single-channel input image of size $28 \times 28$ (see Sec.~\ref{sec:exp}), as the first layer the encoder uses convolution filters of size $4\times 4$ with stride $2$ and $64$ output channels followed by a leaky ReLU activation function. The resulting activation maps go into another convolutional layer with a $128$ channel output using the same filter and stride as well as batch normalization and leaky ReLU. Subsequently the results are fed into a fully connected $1024$-dimensional layer followed by batch normalization and leaky ReLU. The latent variational parameter vector is chosen to be $62$-dimensional and connects linearly with the layer before.
After $z$ is sampled, the decoder maps it to a fully connected layer with output dimension $1024$ followed by a batch normalization layer and a ReLU activation function. It is followed by another fully connected layer with output dimensions $7\times 7 \times 128$, also using batch normalization and ReLU.
Then a deconvolution (or transpose convolution) operation is applied resulting in a $14 \times 14 \times 64$ dimensional output, again followed by batch normalization and ReLU.
Finally, another deconvolution produces a $28 \times 28 \times 1$ output that is followed by sigmoid activation function.

\subsection{Bayesian Neural Network}

Suppose that the predictive function of a classification neural network is solely determined by its weight vector $w$.
If based on training data $X$ and $Y$ we got a posterior distribution over the weights $p(w \mid X,Y) = \frac{p(Y \mid X, w) p(w)}{\int_{w^{'}} p(Y \mid X, w^{'}) p(w^{'}) dw^{'}}$, then for a new input $x^{*}$ we have that
\begin{align}
	p(y^{*} \mid x^{*};X,Y)= \int_w p(y \mid x^{*}, w) p(w \mid X,Y)dw.
\end{align}
To deal with the intractable posterior distribution $p(w \mid X,Y)$, a variational posterior $q_{\phi}(w)$, characterized by a parameter vector $\phi$, is used to approximate it. We have that
\begin{align}
	\log p(y|x) &= E_{q(w)} \log p(y \mid w, x) - D_{KL}(q_{\phi}(w)||p(w)) \nonumber\\ &\quad+D_{KL}(q_{\phi}(w)||p(w \mid x,y)) \nonumber\\
	&=ELBO(\phi) + D_{KL}(q_{\phi}(w)||p(w \mid x,y)),
\end{align}
where we define
\begin{align}
ELBO(\phi) = E_{q(w)} \log p(y \mid w, x) - D_{KL}(q_{\phi}(w)||p(w)).\label{eqn:elbo}
\end{align}
Optimization of $ELBO(\phi)$ with respect to $\phi$ is equivalent to optimizing the KL divergence between the variational posterior $q_{\phi}(w)$ and the posterior $p(w\mid x,y)$.

The variational parameter $\phi$ characterizes a Gaussian distribution on the weights by $q_\phi(w_{ij}) = \mathcal{N}(\mu_{ij}, \sigma^2_{ij})$, where $i$ is the input index and $j$ is the output index. With more complicated index conventions the following analysis can be generalized to convolution operations as well.

\subsubsection{Local reparameterization}

Suppose that the weights $W$ connecting two layers follow an isotropic Gaussian distribution, i.e., $q(w_{ij}) = \mathcal{N}(\mu_{ij}, \sigma^2_{ij})$ as mentioned above. Given an input $A$ the pre-activations $B = AW$ have elements $b_{mj} = \sum_{i}a_{mi}w_{ij}$, where $b_{mj}$ is the $j$th output for the $m$th instance in the minibatch. Each $b_{mj}$ follows a Gaussian distribution with mean and variance given by
\begin{align}
  E(b_{mj}) &= \sum_{i}a_{mi}\mu_{ij}, \label{eqn:w2a1} \\
  var(b_{mj}) &= \sum_{i}a_{mi}^2\sigma_{ij}^2, \label{eqn:w2a2}
\end{align}
So instead of sampling noise variables to reparameterize the weights, a computational graph could directly connect variational parameter of the weight to pre-activations, and sample pre-activations by $b_{mj} = \sum_{i}a_{mi}\mu_{ij} + \zeta_m  \sqrt{\sum_{i}a_{mi}^2\sigma_{ij}^2}$, where $\zeta_m \sim \mathcal{N}(0,1)$.
As pointed out in \cite{Kingma2015-qg} the minibatch variance of the ELBO estimator depends on the covariance of the ELBO estimator across instances in a minibatch. Local reparameterization on the pre-activations, separately for each instance, makes the covariance zero, which could reduce the variance of the ELBO estimator, compared to global reparamerization on the weights. Further arguments on how local reparameterization reduces the variance on the gradients of ELBO with respect to $\sigma_{ij}$ are presented in \cite{Kingma2015-qg}.

\subsubsection{Connection with dropout and variational dropout}

With dropout the pre-activations are given by $B=(A \odot \frac{D^p}{1-p}) W$, where $\odot$ represent element wise product, $D^p$ is the corresponding dropout variable, and $p$ is the dropout rate of input. Suppose that $d^p_{mi}$ is the dropout variable for the $m$th instance, corresponding to input position $i$, then $b_{mj} = \sum_{i}a_{mi}\frac{d^p_{mi}}{1-p}w_{ij}$. If $d^p_{mi} \sim Bernoulli(1-p)$, then
$E(b_{mj}\mid W) = \sum_{i}a_{mi}w_{ij}$ and $var(b_{mj}\mid W)=\sum_{i}a_{mi}^2w_{ij}^2\frac{p}{1-p}$
Thus, approximately $b_{mj}\mid W \sim \mathcal{N}(\sum_{i}a_{mi}w_{ij},\sum_{i}a_{mi}^2w_{ij}^2\frac{p}{1-p})$, which is equivalent to a Gaussian dropout with dropout noise $\mathcal{N}(1, \alpha)$ and $\alpha = \frac{p}{1-p}$.
Comparing with Equation (\ref{eqn:w2a1}) and (\ref{eqn:w2a2}), this is equivalent to parametrizing the variational distribution of the weights as $q(w_{ij})= \mathcal{N}(\mu_{ij},\alpha\mu_{ij}^2)$, where $\sigma_{ij}$ is replaced with $\alpha\mu_{ij}^2$.
To make it fully consistent with Gaussian Dropout, the prior $p(w)$ in Equation (\ref{eqn:elbo}) has to be an improper log-uniform prior so that the second term in Equation(\ref{eqn:elbo}) does not depend on $\mu_{ij}$. However, in this paper, we use a Gaussian proper prior to make it a more general Bayesian neural network. For the calculation of KL divergence instead of using an approximation formula for the improper prior, we simply use the variational Monte Carlo samples as in \cite{blundell2015weight} to calculate the KL divergence as an expectation problem of the log of the likelihood ratio. In terms of architecture, we replace the ReLU to be the softplus activation function in the original CNN architecture.

\section{Related Work}\label{sec:relate}
\textbf{Transfer Learning, Domain Adaptation, Domain Generalization.} 
As mentioned in the introduction part, many domain adaptation and domain generalization benchmark datasets are curated or combination of multiple datasets. Our proposed technique, however, offers possibility to create alternative benchmark datasets, based on only one dataset.

\textbf{Robustness and Generalization of Neural Networks.}
Many attention has been paid to robustness of neural networks recently.  For example, Adversarial examples have been created by distorting clean images slightly and confuse a classifier. Adversarial robustness measures the worst case performance, while corruption robustness measures the classifier's average performance on image corruptions and perturbation robustness measures the prediction stability and consistency under perturbations \cite{hendrycks2019benchmarking}. Benchmark datasets have been created \cite{hendrycks2019benchmarking} for robustness of neural networks under Corruptions and Perturbations.
%
In terms of distribution shift robustness,
where models may silently fail on out-of-distribution samples, it is beneficial to predict out-of-distribution at test time \cite{liang2017enhancing}. In \cite{nalisnick2018deep}, it is reported that out-of-distribution dataset was assigned higher confidence, when training flow based generative models. In the above works, however, distribution shift arises from either changing data or the use of multiple datasets, while, in the proposed resampling technique, we use only one dataset to deliberately generate distribution shifted splits, and the data itself is left intact. 

In terms of generalization, measures like the margin distribution as a predictor for generalization gap is studied \cite{margin}, which we find interesting to evaluate similar measures on splits of train test data created by our resampling technique. 

\textbf{Disentanglement.} Disentanglement tries to find latent representation of data that aligns with independent data generalization factors for interpretation and robust classification \cite{higgins2017beta}. Our method does not aim at achieving disentanglement, i.e., the inferred subdomain labels do not necessarily correspond to any independent data generalization factors. However, our method could potentially testify if disentangled representations can help to overcome model performance deterioration caused by distribution shift.

\section{Experiments}\label{sec:exp}

We use the Fashion-MNIST \cite{xiao2017fashion} data for the initial examples.
Fashion-MNIST consists of $70,000$ grayscale fashion product images of size $28 \times 28$ pixels, which fall into 10 classes ($7000$ images per class).
The original $60,000$ training and $10,000$ test images are combined before the application of cross validation and our resampling method discussed below. Source code and further datasets can be found at \url{https://github.com/compstat-lmu/paper_2019_variationalResampleDistributionShift}.

We compare our resampling method with 5-fold cross validation to demonstrate empirically that distribution shift is indeed  a problem, and we investigate how different CNN models are affected by distribution shift.

\subsection{Data Splitting with Distribution Shift}
\begin{figure}[htb]
\vskip 0.2in
\begin{center}
\centerline{\includegraphics[width=0.9\columnwidth]{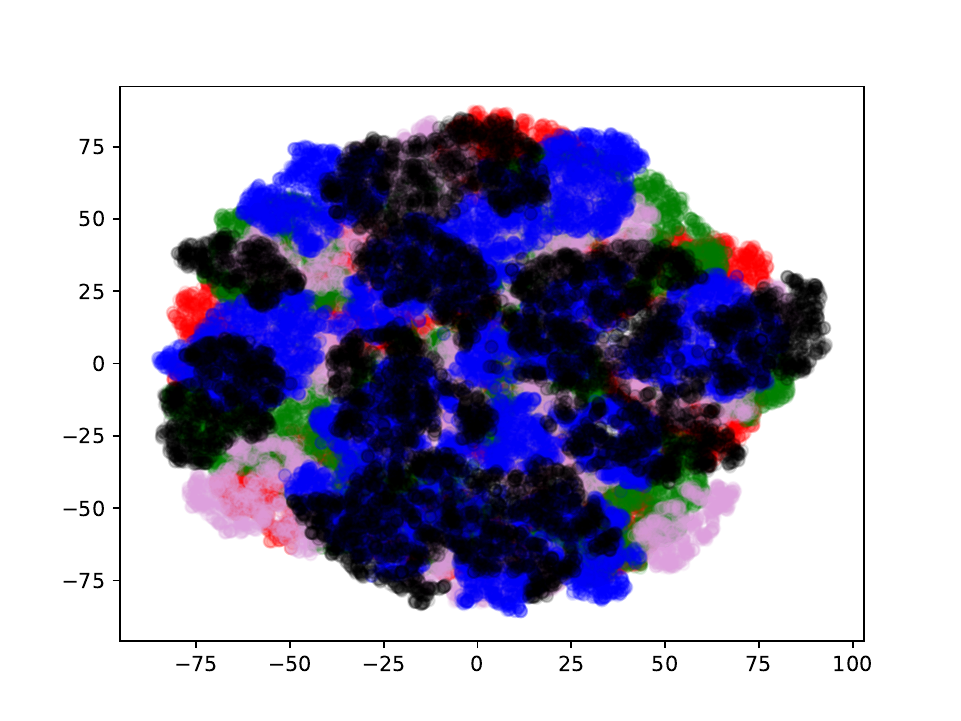}}
\caption{Scatter plot of t-SNE transformed 2-d values from joint data latent representations. Colors indicate different clusters from VGMM-VAE-CV.}
\label{fig:T-SNE}
\end{center}
\vskip -0.2in
\end{figure}

We purposefully obtain data splits with distribution shift using the transformed latent representation from the trained VAE model by methods described in Section \ref{sec:methods} (see Algorithm \ref{alg:dscv-vae}).
In order to visualize the distribution shift we apply the t-SNE algorithm to the total data from all classes latent representation by training a separate VAE, then color data from each cluster from our method to represent the split, as shown in Figure \ref{fig:T-SNE}.
As a quantitative assessment we calculate the Wasserstein distances shown in Section \ref{app:wasser}, from which it is apparent that distances between the splits resulting from our resampling technique are much larger compared to random splitting.

\subsection{Assessment of Performance Deterioration of CNN Models due to Distribution Shift}\label{sec:nonBayesian_experiments}

For the first experiment we use the well known AlexNet \cite{Krizhevsky2012-la} and LeNet \cite{Lecun1998-rs} CNN architectures to perform an image classification task on the Fashion-MNIST data as described in Section \ref{sec:methods}. We also use a simple neural network with three convolutional and three fully connected layers, denoted by 3conv3fc.

The goal of this experiment is two-fold.
(1) We show that we can indeed deliberately subsample a given dataset to create several subsets, which are affected by distribution shift with respect to $p(x|y)$ but roughly share a common distribution $p(y)$ among the clusters (we confirmed this post-hoc empirically).
(2) We demonstrate that distribution shift between the training and the test data substantially reduces the classification accuracy of CNN models on the test data, and it furthermore largely increases the variability in the reported accuracy values.

Both the conventional 5-fold cross validation and the approach of Algorithm \ref{alg:dscv-vae} yield five (train-validation)-test configurations each, where validation takes 20 percent of the train-validation splits randomly.
Thus, each considered CNN is trained and tested five times using conventional cross-validation for data splitting, and five times using our proposed approach.
All models are trained for $100$ epochs, and we record the training, validation, and test accuracies.
Figures \ref{fig:AlexNet}a, \ref{fig:LeNet}a, and \ref{fig:3conv3fc}a show line plots of accuracy by epoch for AlexNet, LeNet, and 3conv3fc respectively, which are separated into individual panels according to the data split (training, validation, test) and data splitting procedure (conventional cross-validation and Algorithm \ref{alg:dscv-vae}).
In particular, the first row of the panels in each figure shows the results from using conventional cross validation for data splitting (i.e., no distribution shift), where we see that the test accuracy is almost identical to the validation accuracy (as one would expect).
The second row of the panels in each figure, however, shows that the test accuracy curves behave wildly different than the validation accuracy curves.
Specifically, the test accuracy is on average substantially reduced when the data are split according to Algorithm \ref{alg:dscv-vae}, i.e., when there is a shift in the conditional feature distribution $p(x|y)$ between the training and the test splits.
Furthermore, we see that distribution shift in $p(x|y)$ also leads to a large increase in the variance of the obtained test accuracy values.
These results are also summarized in Table~\ref{table:vgmm_acc}.

Thus, the comparison between conventional cross validation results (where all training and test distributions are equal) and our resampling approach for data splitting clearly shows that a shift in the conditional feature distributions $p(x|y)$ can lead to a massive deterioration in test data performance, even when the label distributions $p(y)$ are equal.

\subsection{Bayesian CNNs under Distribution Shift}\label{sec:Bayesian_experiments}
The Bayesian approach to deep learning uses distributions over parameters instead of point estimates to represent the model.
This makes Bayesian deep neural networks more robust to overfitting \cite{shridhar2019comprehensive}, and suggests that they may be less affected by distribution shift.
In our second experiment we investigate whether Bayesian CNNs are more robust to distribution shift introduced by our proposed resampling strategy, compared to the conventional CNNs considered in Section \ref{sec:nonBayesian_experiments}.

We use the Bayesian counterparts of the same CNN architectures as considered in Section \ref{sec:nonBayesian_experiments}, such as the Bayesian versions of AlexNet \cite{Krizhevsky2012-la} and LeNet \cite{Lecun1998-rs} introduced by \cite{shridhar2019comprehensive}.
Apart from the substitution of the Bayesian CNN models in place of the frequentist CNN architectures, the experiments are identical to those described in Section \ref{sec:nonBayesian_experiments}. 
Figures \ref{fig:AlexNet}b, \ref{fig:LeNet}b, and \ref{fig:3conv3fc}b as well as Table \ref{table:vgmm_acc} show the results in the same format as described in Section \ref{sec:nonBayesian_experiments}.
While it is apparent from Figures \ref{fig:AlexNet}b and \ref{fig:LeNet}b that Bayesian CNNs are less prone to overfitting to the training data than their conventional CNN counterparts, their vulnerability with respect to distribution shift seems to be about the same. 

Although the Bayesian Neural Network is trained with respect to the variational free energy objective, which shows better generalization to data from the same distribution compared to the frequentist approach \cite{blundell2015weight}, the gradients with respect to the variational parameters are still only based on the training data distribution. In future work, it would also be interesting to investigate if the expressive power of the variational distribution on the weights would be a potential factor to improve.

\subsection{Comparison of CNN models with respect to their robustness to distribution shift}\label{sec:comparison_of_CNNs}

\begin{figure*}[htb]
\begin{center}
\subfigure[AlexNet]{
    \includegraphics[width=0.8\textwidth]{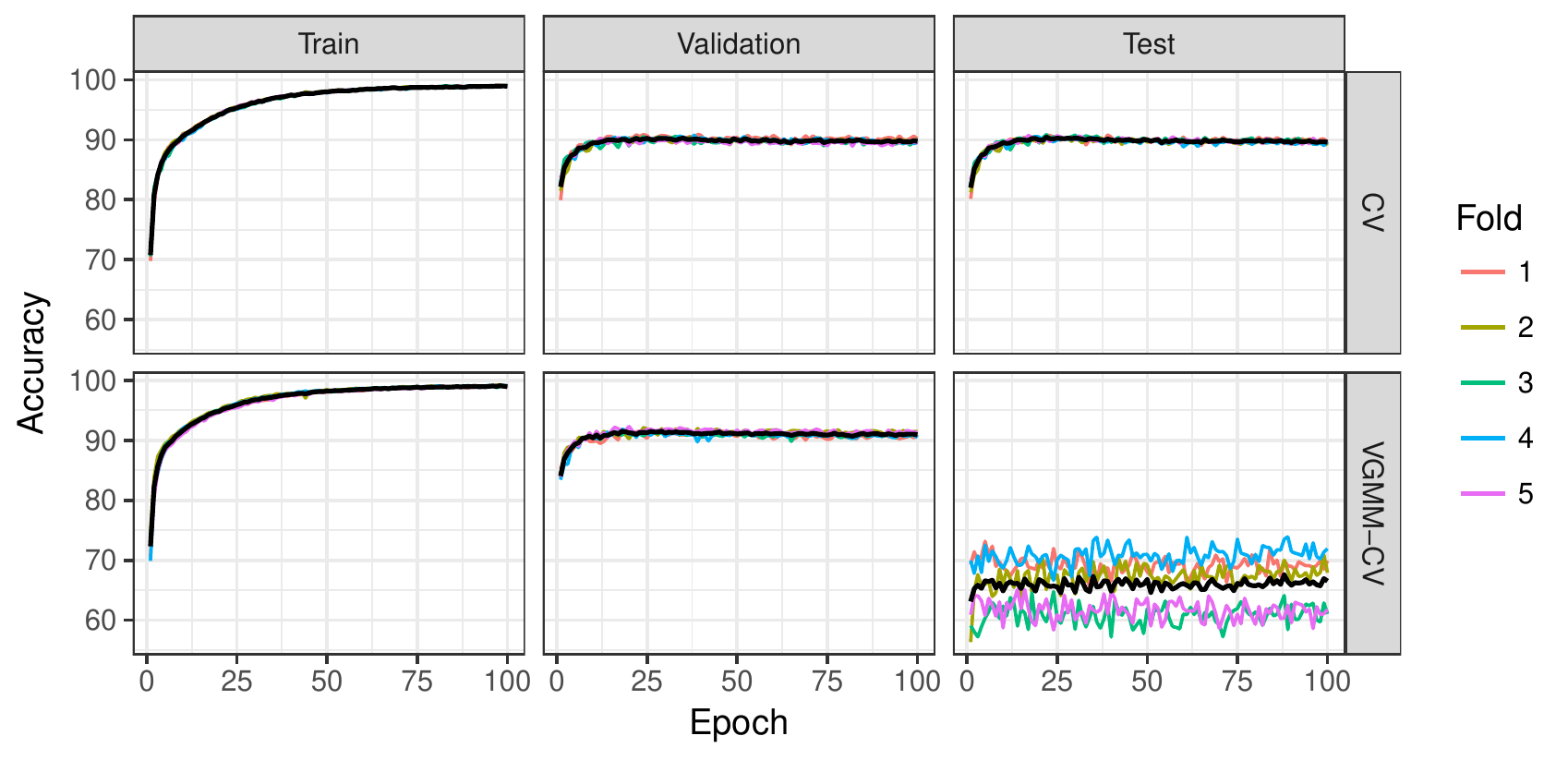}
}
\subfigure[Bayesian AlexNet]{
    \centerline{\includegraphics[width=0.8\textwidth]{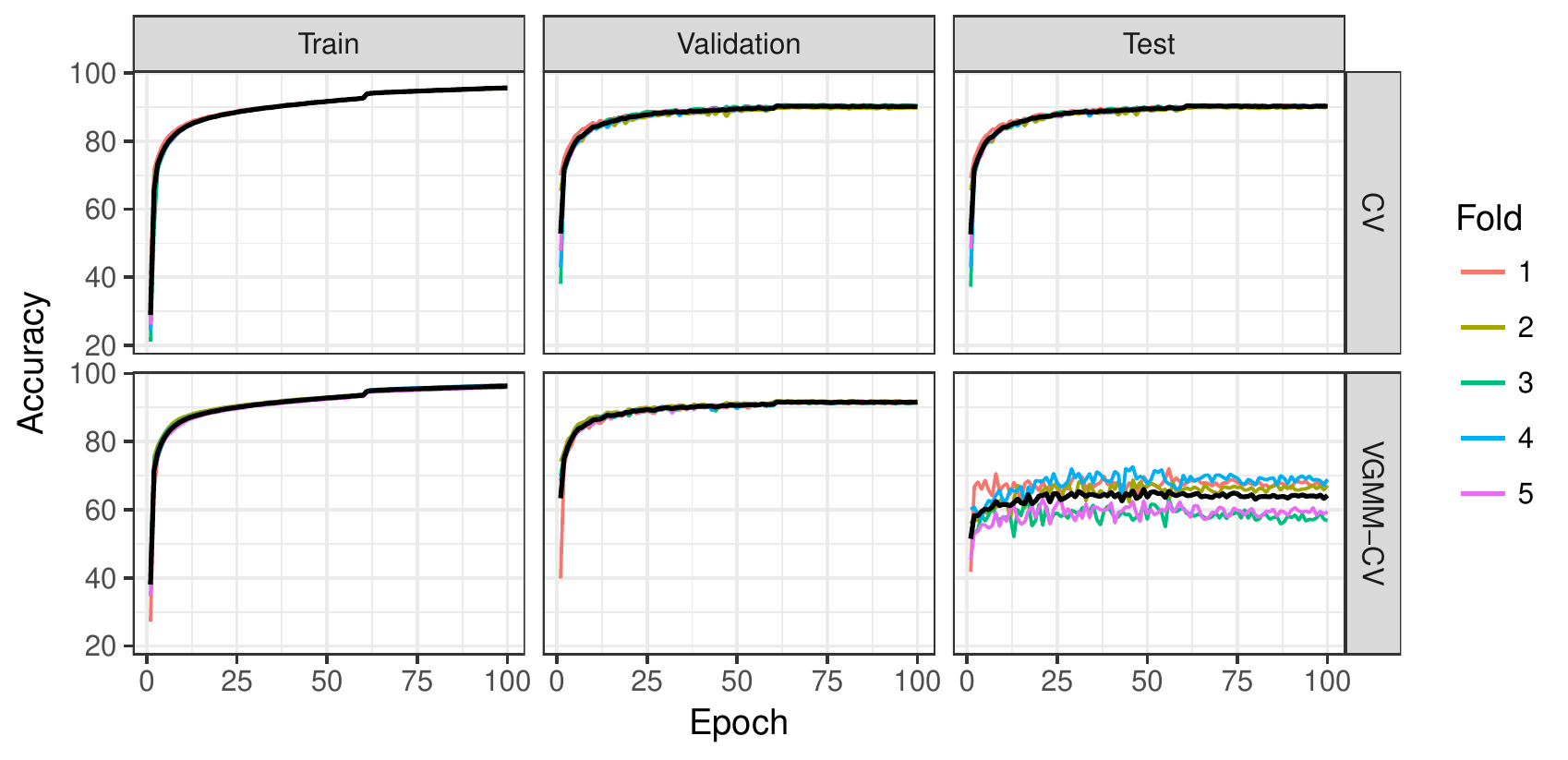}}
}
\caption{(a) AlexNet \cite{Krizhevsky2012-la} and (b) Bayesian AlexNet \cite{shridhar2019comprehensive} classification accuracies on the Fashion-MNIST data are shown by epoch ($1$-$100$), data split (training, validation, test), and data splitting procedure (``CV'' and ``VGMM-CV''). In the first row of panels, entitled ``CV'', the data are split randomly (conventional cross validation). In the second row of panels, entitled ``VGMM-CV'', the data are split as in Algorithm \ref{alg:dscv-vae} leading to a conditional shift in $p(x|y)$ between the training and the test splits. The thicker black line represents the average value across the data splits. We see that a shift in the conditional feature distribution $p(x|y)$ of the test data leads to a reduced accuracy as well as an increased variance.}
\label{fig:AlexNet}
\end{center}
\end{figure*}

\begin{figure*}[htb]
\begin{center}
\subfigure[LeNet]{
    \includegraphics[width=0.8\textwidth]{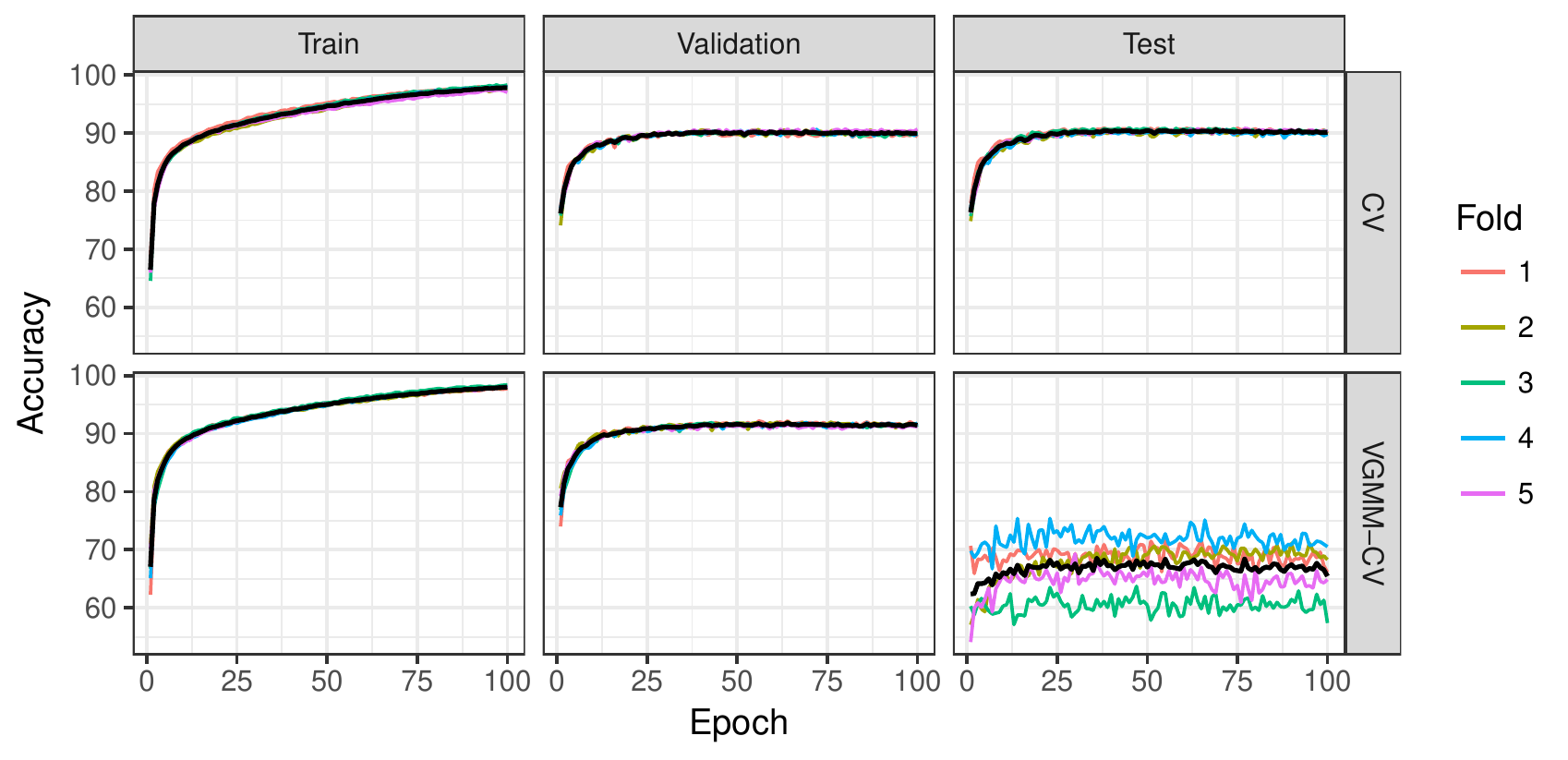}
}
\subfigure[Bayesian LeNet]{
    \centerline{\includegraphics[width=0.8\textwidth]{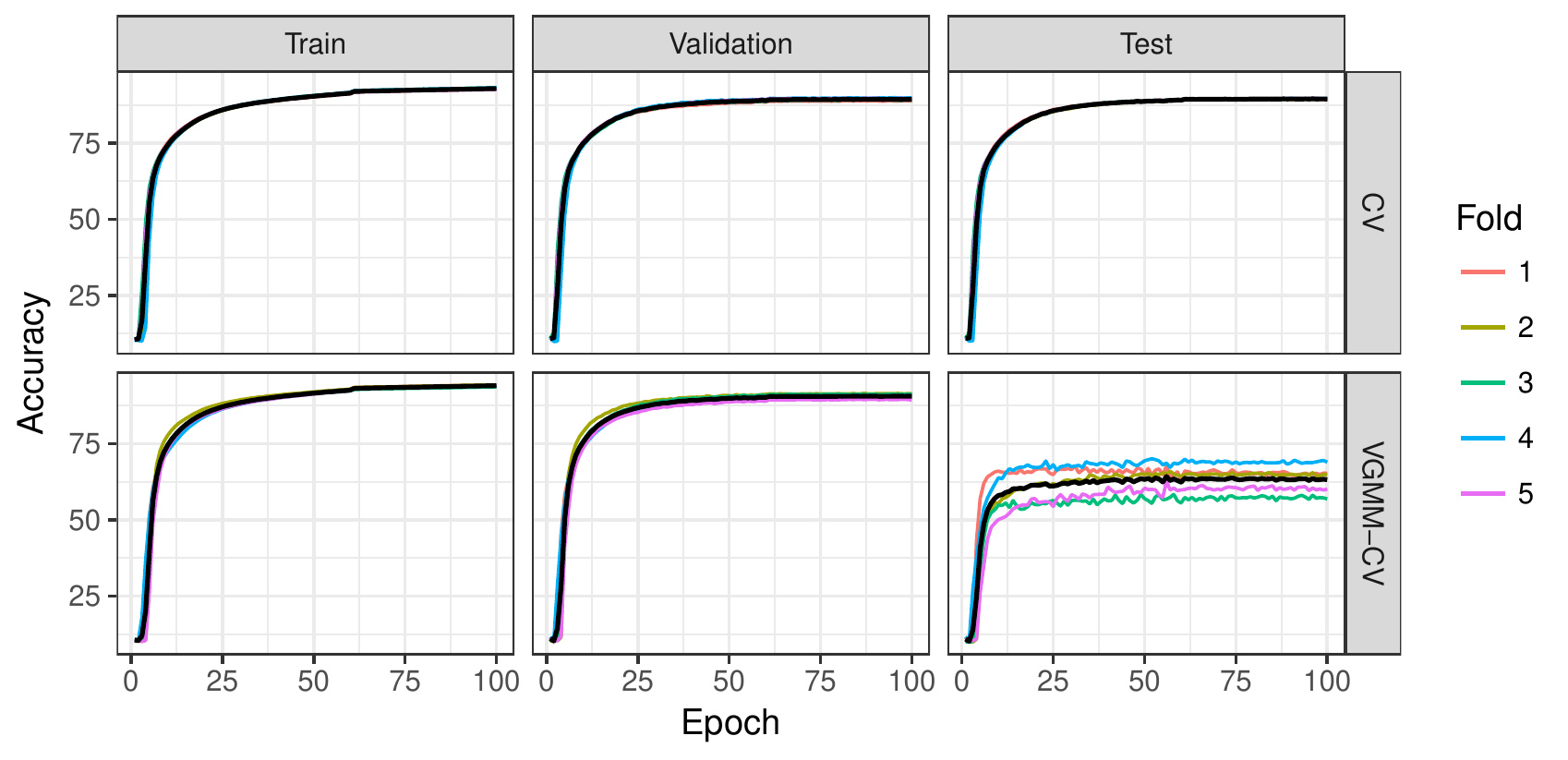}}
}
\caption{(a) LeNet \cite{Lecun1998-rs} and (b) Bayesian LeNet \cite{shridhar2019comprehensive} classification accuracies on the Fashion-MNIST data are shown by epoch ($1$-$100$), data split (training, validation, test), and data splitting procedure (``CV'' and ``VGMM-CV''). In the first row of panels, entitled ``CV'', the data are split randomly (conventional cross validation). In the second row of panels, entitled ``VGMM-CV'', the data are split as in Algorithm \ref{alg:dscv-vae} leading to a conditional shift in $p(x|y)$ between the training and the test splits. The thicker black line represents the average value across the data splits. We see that a shift in the conditional feature distribution $p(x|y)$ of the test data leads to a reduced accuracy as well as an increased variance.}
\label{fig:LeNet}
\end{center}
\end{figure*}

\begin{figure*}[htb]
\begin{center}
\subfigure[3conv3fc]{
    \includegraphics[width=0.8\textwidth]{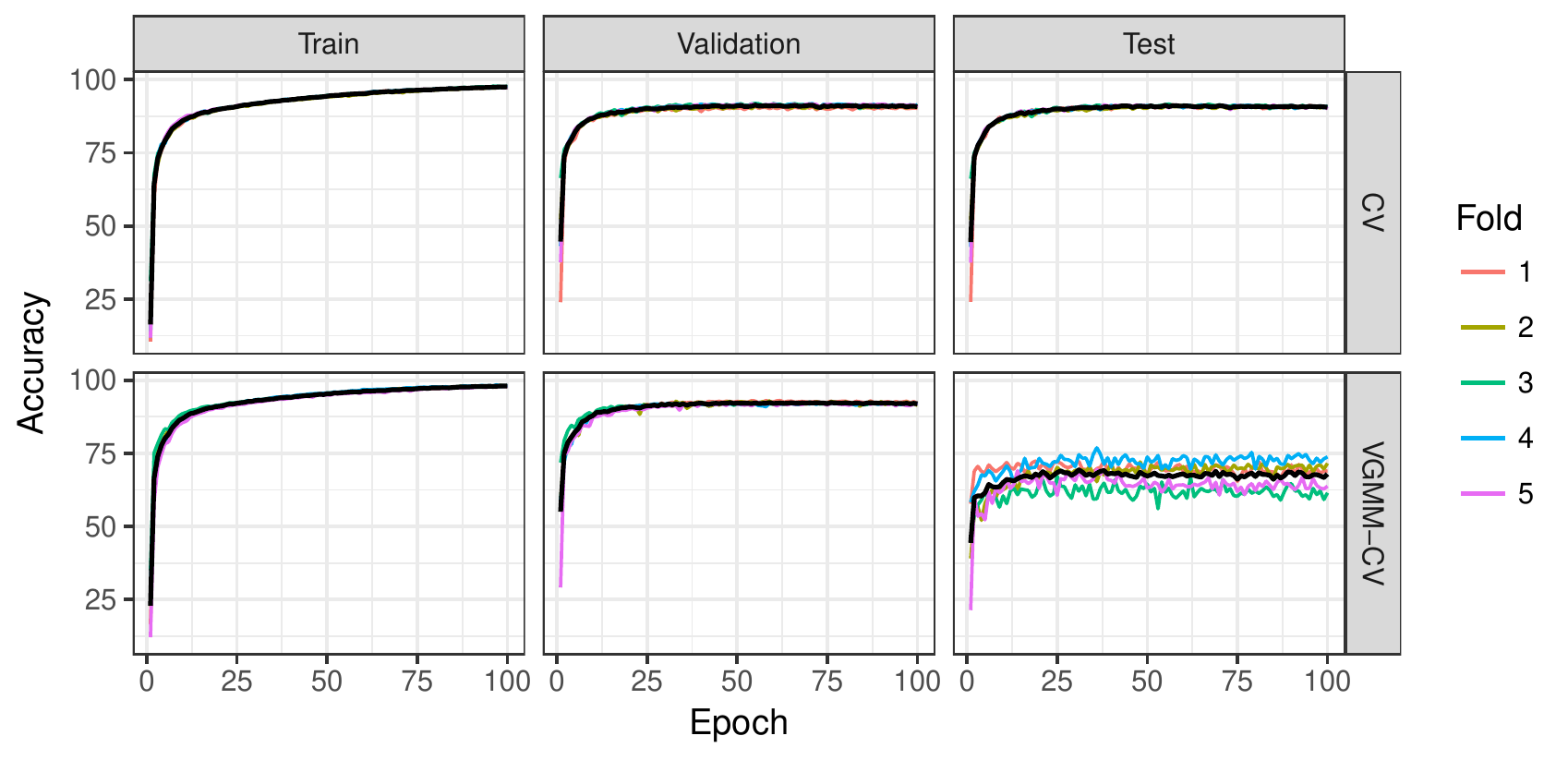}
}
\subfigure[Bayesian 3conv3fc]{
    \centerline{\includegraphics[width=0.8\textwidth]{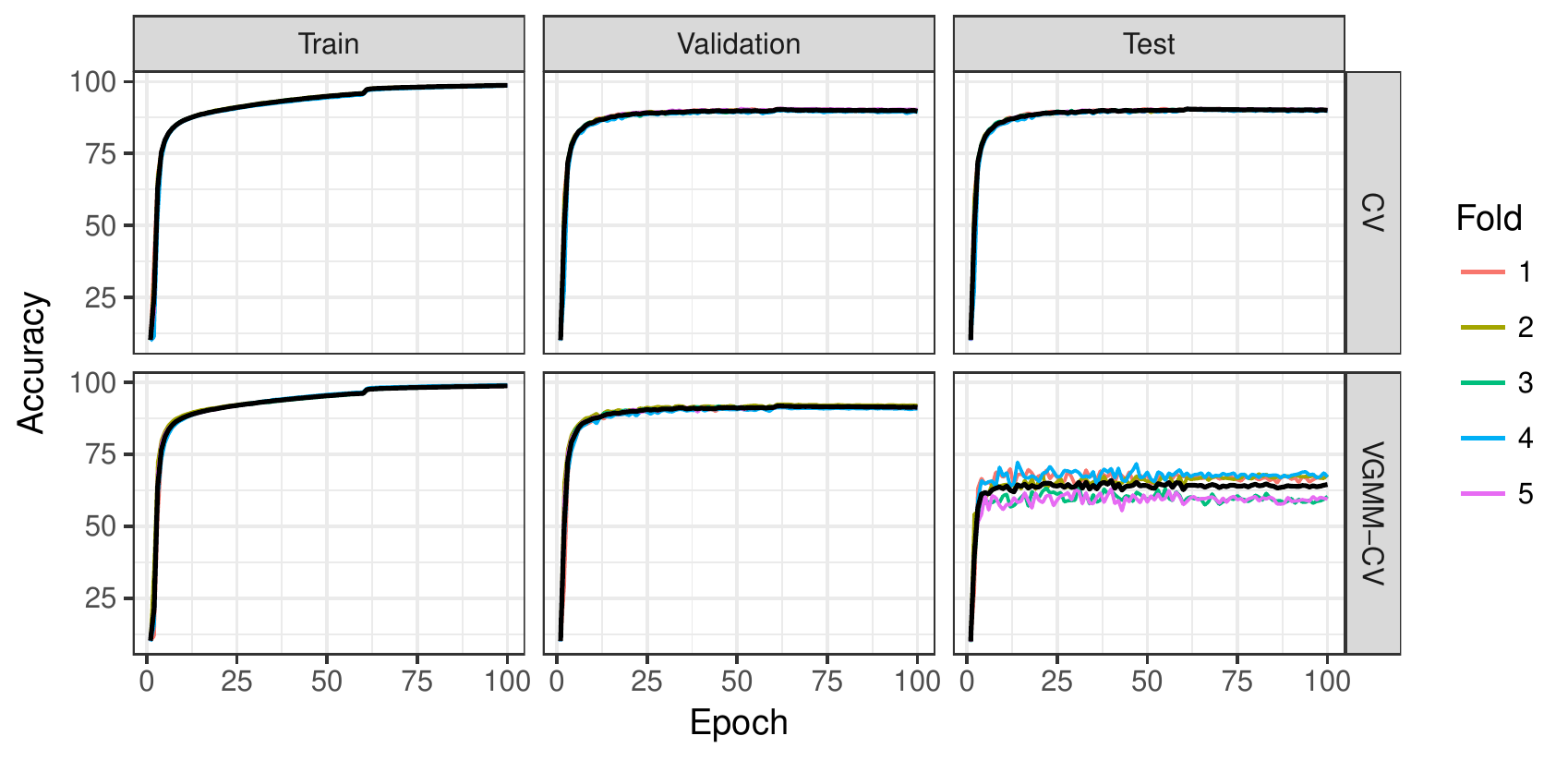}}
}
\caption{(a) A simple neural network with three convolution and three fully connected layers (3conv3fc) and (b) Bayesian 3conv3fc classification accuracies on the Fashion-MNIST data are shown by epoch ($1$-$100$), data split (training, validation, test), and data splitting procedure (``CV'' and ``VGMM-CV''). In the first row of panels, entitled ``CV'', the data are split randomly (conventional cross validation). In the second row of panels, entitled ``VGMM-CV'', the data are split as in Algorithm \ref{alg:dscv-vae} leading to a conditional shift in $p(x|y)$ between the training and the test splits. The thicker black line represents the average value across the data splits. We see that a shift in the conditional feature distribution $p(x|y)$ of the test data leads to a reduced accuracy as well as an increased variance.}
\label{fig:3conv3fc}
\end{center}
\end{figure*}

Because under our proposed resampling approach the validation and the training data share the same distribution but the conditional feature distribution $p(x|y)$ of the test data is shifted, the robustness of a CNN to distribution shift can be quantified by comparing the test accuracy curves to the validation accuracy curves in our experiments (see Figures \ref{fig:AlexNet}, \ref{fig:LeNet}, and \ref{fig:3conv3fc}).
There are different approaches to carry out such a comparison.
However, for simplicity in this work we compare only the empirical mean and standard deviation values at the last epoch.
Table~\ref{table:vgmm_acc} summarizes these values.
We see that the classification accuracy on the test data reduces by about $26.0$ points on average due to distribution shift.
In addition, in the presence of distribution shift the standard deviation of the reported test accuracy values is about 14 times larger than the standard deviation of the accuracy values on the validation data.

While the degree of performance deterioration as measured by these analyses seems to be about the same between all considered CNN models, it is conceivable that some models will be more or less affected by distribution shift, which will be reflected in the values and accuracy curves as analyzed above.
Hence, our framework provides a way to quantitatively compare the robustness to distribution shift between different models.

As an additional point of reference, Table \ref{table:original_train-test_acc} contains the classification accuracies after 100 training epochs for the same CNN architectures on the original train-test split provided in the Fashion-MNIST \cite{xiao2017fashion} data. Note that there is no distribution shift between the training and the testing data in this case, and the training dataset is larger than in the experiments of Sections \ref{sec:nonBayesian_experiments} and \ref{sec:Bayesian_experiments}.

\begin{table}[htb]
\vskip 0.15in
\begin{center}
\begin{small}
\begin{sc}
\begin{tabular}{lcccr}
\toprule
Model & Train. acc. & Val. acc. & Test acc \\
\midrule
\midrule
AlexNet  & 98.97 (0.13) & 90.93 (0.43) & 66.51 (4.88) \\
\midrule
\specialcell{Bayesian\\AlexNet}  & 96.33 (0.29) & 91.58 (0.25) & 64.21 (5.67) \\
\midrule
LeNet     & 98.03 (0.25) & 91.44 (0.29) & 65.43 (5.00) \\
\midrule
\specialcell{Bayesian\\LeNet}    & 94.04 (0.27) & 90.54 (0.82) & 63.18 (4.72) \\
\midrule
3conv3fc  & 97.97 (0.13) & 91.91 (0.28) & 67.92 (5.18) \\
\midrule
\specialcell{Bayesian\\3conv3fc}  & 98.69 (0.19) & 91.26 (0.43) & 64.44 (4.19) \\
\bottomrule
\end{tabular}
\end{sc}
\end{small}
\end{center}
\caption{Average classification accuracies on the training, validation, and testing data splits after 100 training epochs for several CNN models. Empirical mean values with standard deviation in parentheses are computed across the five data splits which are obtained by Algorithm \ref{alg:dscv-vae}.}
\label{table:vgmm_acc}
\vskip -0.1in
\end{table}   

\begin{table}[htb]
\vskip 0.15in
\begin{center}
\begin{small}
\begin{sc}
\begin{tabular}{lcccr}
\toprule
Model & Train. acc. & Test acc \\
\midrule
\midrule
AlexNet    & 98.78 & 89.42 \\
\midrule
\specialcell{Bayesian\\AlexNet}    & 95.86 & 90.56 \\
\midrule
LeNet & 97.71 & 89.77 \\
\midrule
\specialcell{Bayesian\\LeNet} & 93.38 & 89.06 \\
\midrule
3conv3fc & 97.37 & 91.08 \\
\midrule
\specialcell{Bayesian\\3conv3fc} & 98.74 & 90.58 \\
\bottomrule
\end{tabular}
\end{sc}
\end{small}
\end{center}
\caption{Classification accuracy after 100 training epochs of multiple CNN models on the original train-test split provided in the Fashion-MNIST \cite{xiao2017fashion} data.}
\label{table:original_train-test_acc}
\vskip -0.1in
\end{table}

\subsection{Computed pairwise Wasserstein distances}
\label{app:wasser}
With the python package \texttt{POT} \cite{flamary2017pot} one can compute the pairwise Wasserstein distance between two clusters of data. In Table \ref{tb:eq:wasserstein_vgmm} we computed the pairwise Wasserstein distances across the 5 clusters created based on VGMM as described in Section \ref{sec:methods}, which correspond to a conditional distribution shift in $p(x|y)$. In Table \ref{tb:eq:wasserstein_random} the pairwise Wassserstein distances are computed based on random splits as in conventional cross validation. It can be clearly seen that the VGMM variant creates larger pairwise Wasserstein distances, which testifies that our proposed method generates splits of data with significant distribution shift, as intended.  
\begin{table}
\caption{Pairwise Wasserstein distance across 5 clusters created by Varitional Gaussian Mixture Models on the data latent space}
\begin{equation*}
\footnotesize{
\begin{bmatrix}
$0.0$        & $0.22072112$ & $0.21187810$ & $0.22447902$ & $0.21012454$ \\
$0.22068973$ & $0.0$        & $0.22250834$ & $0.21232671$ & $0.20131575$ \\
$0.21465105$ & $0.22328222$ & $0.0$        & $0.23423279$ & $0.21377970$ \\
$0.23307045$ & $0.20407709$ & $0.23252131$ & $0.0$        & $0.20784507$ \\
$0.20973221$ & $0.20124379$ & $0.21279558$ & $0.20822021$ & $0.0$
\end{bmatrix}
}
\end{equation*}
\label{tb:eq:wasserstein_vgmm}
\end{table}

\begin{table}
	\caption{Pairwise Wasserstein distance across 5 clusters created by random splitting of the data}
\begin{equation*}
\footnotesize{
\begin{bmatrix} 
$0.0$        & $0.13235691$ & $0.14389180$ & $0.14445968$ & $0.13845333$ \\
$0.13221454$ & $0.0$        & $0.13702394$ & $0.13594290$ & $0.13507776$ \\
$0.14411578$ & $0.13698794$ & $0.0$        & $0.14563888$ & $0.14467288$ \\
$0.14428186$ & $0.13616575$ & $0.14567640$ & $0.0$        & $0.13870558$ \\
$0.13836885$ & $0.13548402$ & $0.14428162$ & $0.13905402$ & $0.0$
\end{bmatrix}
}
\end{equation*}
\label{tb:eq:wasserstein_random}
\end{table}

\section{Summary and Conclusion}\label{sec:summary}

We propose a new resampling technique to create pseudo subdomains over one dataset.
Our resampling strategy purposefully identifies data splits with distribution shift with respect to the conditional distribution $p(x|y)$ of features $x$ given the label $y$ by utilizing the latent representation of data through generative models. Variational methods are used to assign instances to the pseudo subdomains, which are represented as clusters in the latent space.

We use our new resampling technique to assess the robustness of deep neural networks in terms of generalization ability to distribution shift. We show that CNN models display substantial reductions in performance and an increase in variability under the proposed resampling technique compared to conventional cross validation. This demonstrates the severe problem that the performance of CNN models is strongly affected by changes in the conditional distribution $p(x|y)$ even when the label distribution $p(y)$ remains unchanged and all data originate from the same domain. In addition, we observe that this problem persists for Bayesian CNNs considered in this work, even though Bayesian CNNs otherwise are known to possess superior generalization properties at least for data from the same distribution. Possibly since the gradients with respect to the variational parameters are also based on data from the training distribution, it makes it difficult to generalize to another distribution.

Our approach can be used for the evaluation of the generalization ability of deep learning models and inform model selection, alongside conventional performance evaluation approaches such as cross validation and testing on holdout data. For instance, Automatic Machine Learning (AutoML) \cite{sun2019reinbo} methods should also take distribution shift into account when searching for a model, where our method could easily create splits to serve within an objective function to be optimized during the AutoML process.

There remain some open questions and potential drawbacks of our method. It is yet unknown what model architecture or choice of hyperparameters will affect the created subdomains. Additionally, our artificially created pseudo subdomains do not necessarily correspond to real world (sub)domains, and it is not clear to what extent the artificially created distribution shift is comparable to the types of distribution shift observed between different (sub)domains in the real world.

In future work, methods similar to Restrictive Federated Model Selection \cite{sun2019high} could be used to adapt to the distribution shift generated by the methods proposed in this work. In addition, it would be interesting to see new methods that not only create distribution shift, but also allow to control the extent of distribution shift by use of appropriate hyperparameters, as well as methods which could create pseudo subdomains on tasks other than classification, such as recommendation systems \cite{kushwaha2018lesson}. Furthermore, it would be interesting to see how our approach would serve as a benchmark method to evaluate different domain adaptation and domain generalization algorithms.

\bibliographystyle{IEEEtran}
\bibliography{main}

\end{document}